\documentclass{article}

\usepackage[preprint]{spconf}
\usepackage[utf8]{inputenc} 
\usepackage[T1]{fontenc}    
\usepackage{hyperref}       
\usepackage{url}            
\usepackage{booktabs}       
\usepackage{amsfonts}       
\usepackage{nicefrac}       
\usepackage{microtype}      
\usepackage{graphicx} 
\usepackage{algorithm}
\usepackage{algorithmic}
\usepackage{amssymb,amsmath}
\usepackage{color,soul}
\usepackage[caption=false]{subfig}
\usepackage{todonotes}
\usepackage{multirow}
\usepackage{tikz}
\usetikzlibrary{positioning,shapes,arrows,fit}
\newcommand{\E}{\textbf{E}}
\renewcommand{\L}{\mathcal{L}}

\title{Learning Hard Alignments with Variational Inference}

\name{
Dieterich Lawson\sthanks{Equal contribution.},
Chung-Cheng Chiu$^*$,
George Tucker$^*$,
Colin Raffel,
Kevin Swersky,
Navdeep Jaitly}

\address{Google Brain \\ \{dieterichl,chungchengc,gjt,craffel,kswersky,ndjaitly\}@google.com\\}

\begin{document}
\ninept
\maketitle

\begin{abstract}
There has recently been significant interest in hard attention models for tasks such as object recognition, visual captioning and speech recognition. Hard attention can offer benefits over soft attention such as decreased computational cost, but training hard attention models can be difficult because of the discrete latent variables they introduce. Previous work used REINFORCE and Q-learning to approach these issues, but those methods can provide high-variance gradient estimates and be slow to train. In this paper, we tackle the problem of learning hard attention for a sequential task using variational inference methods, specifically the recently introduced VIMCO and NVIL. Furthermore, we propose a novel baseline that adapts VIMCO to this setting. We demonstrate our method on a phoneme recognition task in clean and noisy environments and show that our method outperforms REINFORCE, with the difference being greater for a more complicated task.
\end{abstract}

\section{Introduction}
Attention models have gained widespread traction from their successful use in tasks such as object recognition, machine translation, speech  recognition where they are used to integrate information from different parts of the input before producing outputs. Soft attention does this by weighting and combining all input elements into a context vector while hard attention selects specific inputs and discards others, leading to computational gains and greater interpretability. While soft attention models are differentiable end-to-end and thus easy to train, hard attention models introduce discrete latent variables that often require reinforcement learning style approaches.

Classic reinforcement learning methods such as REINFORCE \cite{policygradients} and Q-learning \cite{qlearning} have been used to train hard attention models, but these methods can provide high-variance gradient estimates, making training slow and providing inferior solutions. An alternative to reinforcement learning is variational inference, which trains a second model, called the approximate posterior, to be close to the true posterior over the latent variables. The approximate posterior uses information about both the input and its labels to produce settings of the latent variables used to train the original model. This can provide lower-variance gradient estimates and better solutions.

In this paper, we leverage recent developments in variational inference to fit hard attention models in a sequential setting. We specialize these method to sequences and develop a model for the approximate posterior. In response to issues applying variational inference techniques to long sequences, we develop new variance control methods. Finally we show experimentally that our approach improves performance and substantially improves training time for speech recognition on the TIMIT dataset as well as a challenging noisy, multi-speaker version of TIMIT that we call Multi-TIMIT.
\section{Methods}

\begin{figure}[t]
\subfloat[Model]{%
\begin{tikzpicture}[
square/.style={rectangle, draw=black, minimum height=8mm, minimum width=8mm},
round/.style={circle, draw=black, minimum size=8mm},
dark_round/.style={circle, draw=black, fill=lightgray, minimum size=8mm},
]
\node [square] (x1) {$x_1$};
\node [square] (x2) [right=0.5cm of x1] {$x_2$};
\node [square] (x3) [right=0.85cm of x2] {$x_2$};

\node [square] (h1) [above=0.25cm of x1] {$h_1$};
\node [square] (h2) [above=0.25cm of x2] {$h_2$};
\node [square] (h3) [above=0.25cm of x3] {$h_3$};

\node [dark_round] (b1) [above=0.25cm of h1] {$b_1$};
\node [] (h2_y1) [above=0.5cm of h2] {};
\node [round] (b2) [right=0.01cm of h2_y1] {$b_2$};
\node [dark_round] (b3) [above=0.25cm of h3] {$b_3$};

\node [square] (y1) [above=0.5cm of h2_y1] {$y_1$};

\draw[->] (x1) -- (h1);
\draw[->] (x2) -- (h2);
\draw[->] (x3) -- (h3);

\draw[->] (h1) -- (h2);
\draw[->] (h2) -- (h3);

\draw[->] (h1) -- (b1);
\draw[->] (h2) to[out=90, in=-90, looseness=0.5] (b2);
\draw[->] (h3) -- (b3);

\draw[->] (h2) -- (y1);

\draw[->] (b2) to[out=90, in=-90, looseness=0.5] (y1);

\draw[->] (b1) to[out=0, in=-180, looseness=0.5] (h2);
\draw[->] (b2) to[out=0, in=-180, looseness=0.5] (h3);

\draw[->] (y1) to[out=0, in=-180, looseness=0.5] (h3);

\end{tikzpicture}
}
\hspace{5mm}
\subfloat[Approximate Posterior]{
\begin{tikzpicture}[
square/.style={rectangle, draw=black, minimum height=8mm, minimum width=8mm},
round/.style={circle, draw=black, minimum size=8mm},
dark_round/.style={circle, draw=black, fill=lightgray, minimum size=8mm}
]

\node [square] (h1) {$h_1$};
\node [square] (h2) [right=0.5cm of h1] {$h_2$};
\node [square] (h3) [right=0.5cm of h2] {$h_3$};

\node [dark_round] (b1) [above=0.25cm of h1] {$b_1$};
\node [round] (b2) [above=0.25cm of h2] {$b_2$};
\node [dark_round] (b3) [above=0.25cm of h3] {$b_3$};

\node [] (h1_h1p) [below=0.5cm of h1] {};
\node [] (h2_h2p) [below=0.5cm of h2] {};
\node [] (h3_h3p) [below=0.5cm of h3] {};

\node [square] (y1) [left=0.01cm of h1_h1p] {$y_1$};
\node [square] (y2) [left=0.01cm of h2_h2p] {$y_1$};
\node [square] (y3) [below=0.225cm of h3] {$y_2$};

\node [square] (h1p) [below=0.5cm of h1_h1p] {$h'_1$};
\node [square] (h2p) [below=0.5cm of h2_h2p] {$h'_2$};
\node [square] (h3p) [below=0.5cm of h3_h3p] {$h'_3$};

\node [square] (x1) [below=0.25cm of h1p] {$x_1$};
\node [square] (x2) [below=0.25cm of h2p] {$x_2$};
\node [square] (x3) [below=0.25cm of h3p] {$x_3$};

\draw[->] (x1) -- (h1p);
\draw[->] (x2) -- (h2p);
\draw[->] (x3) -- (h3p);

\draw[<->] (h1p) -- (h2p);
\draw[<->] (h2p) -- (h3p);

\draw[->] (h1p) -- (h1);
\draw[->] (h2p) -- (h2);
\draw[->] (h2p) to[out=90, in=-90, looseness=0.95] (h3);

\draw[->] (y1) to[out=90, in=-90, looseness=0.5] (h1);
\draw[->] (y2) to[out=90, in=-90, looseness=0.5] (h2);
\draw[->] (y3) to[out=90, in=-90, looseness=0.5] (h3);

\draw[->] (h1) -- (h2);
\draw[->] (h2) -- (h3);

\draw[->] (h1) -- (b1);
\draw[->] (h2) to[out=90, in=-90, looseness=0.5] (b2);
\draw[->] (h3) -- (b3);

\draw[->] (b1) to[out=0, in=-180, looseness=0.5] (h2);
\draw[->] (b2) to[out=0, in=-180, looseness=0.5] (h3);
\end{tikzpicture}
}
\caption{A diagram of our models.  $b$s denote the Bernoulli emission decision variables, $x$s are inputs, $y$s are targets, and $h$s and $h'$s are the hidden states of the recurrent neural networks (RNNs) that parameterize the conditional distributions of the models. Square nodes are deterministic, round nodes are stochastic. A shaded $b_i$ indicates that the model chose to consume an input and not emit an output while an unshaded $b_i$ mean that the model chose to produce an output and not consume an input. For example, in (a) note that $b_1$ is shaded, so the model did not produce an output on timestep 1 and instead consumes the input $x_2$ on the next timestep. $b_2$ is unshaded, so on the second timestep the model produced output $y_1$.}
\label{model-diagram}
\end{figure}
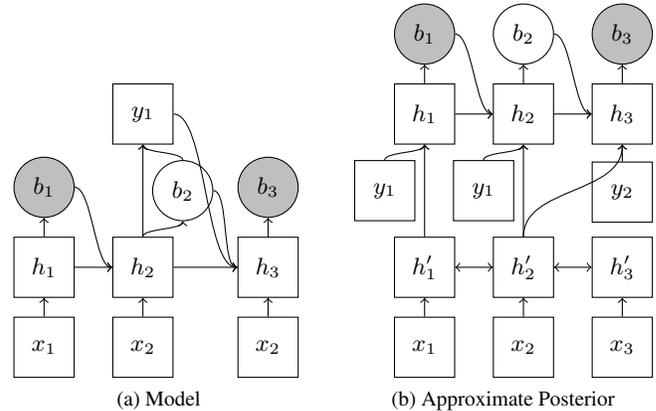 

\subsection{Model}

In this paper we use the online sequence-to-sequence model described in \cite{yuping} to demonstrate our methods. We model $p(y,b|x)$ where $y=y_1,\ldots,y_n$ is a sequence of observed target tokens and $x=x_1,\ldots,x_m$ is a sequence of observed inputs. The Bernoulli latent variables  $b=b_1,\ldots,b_{m+n}$ define when the model outputs tokens, i.e. $b_t=1$ implies the model emitted a token at timestep $t$, and $b_t=0$ implies the model did not emit a token at timestep $t$. If $b_t=1$, the model is forced to dwell on the same input at the next time step, i.e. the observation fed in at timestep $t$ is fed in again at timestep $t+1$ when $b_t=1$. Let $n$ be the number of target tokens, $m$ the number of inputs, and $T=m+n$ the number of steps the model is run for. Our model assumes $p(y,b|x)$ factorizes as
\begin{align}
  \begin{split}
  \label{eq:model}
    p(y,b|x) = \prod_{t=1}^T &p(y_{O(t)}|b_{1:t},x_{1:I(t)},y_{1:O(t-1)})^{b_t}\times \\
    &p(b_t|b_{1:t-1},x_{1:I(t)},y_{1:O(t-1)})
  \end{split}
\end{align}
where $O(t) = \sum_{i=1}^t b_i$ is the position in the output at time $t$ and $I(t) = 1 + \sum_{i=1}^{t-1} (1- b_i)$ is the input position at time t. Intuitively, this expression is the product over time of the probability assigned to the current ground truth given that the model emitted, multiplied by the probability that the model emitted. When $b_t = 0$ the model did not emit at time $t$, so there is no probability assigned to the ground truth on that timestep. For brevity, we will use $y_t$ to implicitly mean $y_{O(t)}$ (i.e., the target at step $t$). Similarly, we will refer to $x_{I(t)}$ as $x_t$ and similarly for ranges over time for these variables.
\subsection{Learning}

To fit the model (\ref{eq:model}) with maximum likelihood we are concerned with maximizing the probability of the observed variables $y$. However, (\ref{eq:model}) is written in terms of the unobserved latents, $b$, so we must marginalize over them. We maximize
\begin{align*}
    \E_b \biggr[  \sum_{t=1}^T b_t \log p(y_t|s_t, b_t)  \biggr] =& \E_{b} \left[\log \prod_{t=1}^T p(y_t|s_t, b_t)^{b_t} \right] \\
    \leq& \log p(y|x),
\end{align*}
where $s_t = \{b_{1:{t-1}}, x_{1:t}, y_{1:{t-1}}\}$ is the state at time $t$ and the expectations are over $p(b_t | s_t)$. Note that this is a lower bound on the log probability of the observed $y$, so maximizing this bound will hopefully increase the likelihood of the observed data. Differentiating this objective gives
\begin{align}
    \textbf{E}_{b} & \biggr[ \nabla \sum_{t=1}^T b_t \log p(y_t| s_t, b_t ) \biggr] + \sum_{t=1}^T \E_b \biggr[ R_t \nabla \log p(b_t |s_t) \biggr]
    \label{eq:reinforce_grad}
\end{align}
where $R_t = \sum_{t' \geq t}^T \log p(y_{t'}| s_{t'}, b_{t'})$ is the return at timestep $t$,  understood intuitively as the log probability the model assigns to observed data for a given series of emission decisions. The first gradient term can be estimated with a single Monte Carlo sample, but the second term exhibits high variance because it involves an unbounded log probability. To reduce variance, \cite{yuping} subtracts a learned baseline $c(b_{1:t-1},x_{1:T}, y_{1:T})$ from the return, which does not change the expectation as long as it is independent of $b_t$.

Performing stochastic gradient ascent with this gradient estimator is the standard REINFORCE algorithm where the reward is the log-likelihood. Unfortunately, this requires sampling $b_t$ from $p(b_t|s_t)$ during training, which can lead to gradient estimates with high variance when settings of $b$ that assign high likelihood to $y$ are rare \cite{variationalimco}. Variational inference is a family of techniques that use importance sampling to instead sample $b$ from a different model, called the approximate posterior or $q$, which approximates the true posterior over $b$, $p(b|x,y)$. We factorize the approximate posterior as
\begin{equation}
 q(b|x,y) = \prod_{t=1}^T q(b_t|b_{1:t-1},x_{1:T}, y_{1:T}).
\end{equation}
The approximate posterior has access to all past and future $x$ and $y$, as well as past $b$, and leverages this information to assign high probability to $b$ that produce large values of $p(y|b,x)$. Intuitively, in speech recognition, knowing the token the model must emit is helpful in deciding when to emit.

Using $q$ and an importance sampling identity we obtain a lower bound on the log-likelihood
\begin{align}
\label{eq:ss_vi_lowerbound}
    \log p(y|x) &= \log\hspace{.05cm} \textbf{E}_{b\sim q}\left[ \frac{p(y,b|x)}{q(b|x,y)}\right] \geq \textbf{E}_{b\sim q}\left[ \log \frac{p(y,b|x)}{q(b|x,y)}\right]
\end{align}
where we can simultaneously optimize $q$ and the parameters of the model to improve the lower bound. Optimizing this bound via stochastic gradient ascent can be thought of as training $p$ with maximum likelihood to reproduce $b$s sampled from $q$. $q$ is then updated with REINFORCE-style gradients where the reward is the log-probability $p$ assigns to $y$ given $b$, similar to (\ref{eq:reinforce_grad}), see \cite{variationalimco} for details. Setting $q(b|x,y) = \prod_t p(b_t|s_t)$ recovers the REINFORCE objective.

\subsubsection{Multi-sample Objectives}
Both the REINFORCE and the variational inference objectives admit multi-sample versions that give tighter bounds on the log-likelihood \cite{iwae}. In particular, the multi-sample variational lower bound is
\begin{align}
\label{eq:ms_vi_lowerbound}
    \L = \textbf{E}_{b^{(1:k)} \sim q} &\left[\log\left( \frac{1}{k} \sum_{i=1}^k \frac{p(y,b^{(i)}|x)}{q(b^{(i)}|x,y)} \right)\right] 
\end{align}
where $k$ is the number of samples and $b^{(i)}$ denotes the $i$th sample of the latent variables. Setting $q(b|x,y) = \prod_t p(b_t|s_t)$ recovers the multi-sample analogue to REINFORCE. 

The gradient of (\ref{eq:ms_vi_lowerbound}) takes a similar form to (\ref{eq:reinforce_grad}), with one low-variance term and one REINFORCE-style term with high variance, for details see \cite{variationalimco}. Similarly to the REINFORCE objective, we can use a baseline $c(b^{(i)}_{1:t-1}, x_{1:T}, y_{1:T}, b^{(-i)}_{1:T})$ to reduce the variance of the gradient as long as it does not depend on $b^{(i)}_t$. Notably, the baseline for trajectory $i$ is allowed to depend on all timesteps of other trajectories, i.e. $b^{(-i)}_{1:T}$.

\subsection{Variance Reduction}
Training these models is challenging due to high variance gradient estimates. We can reduce the variance of the estimators by using information from multiple trajectories to construct baselines. In particular, for REINFORCE, we can write the gradient update as
\begin{equation*}
    \E_{b^{(i)}} \left[ \sum_{t=1}^T \left(R_t - c(s^{(i)}_{t-1}, \{ R_{1:T}^{(j)} \}_{j\neq i})\right) \nabla \log p(b^{(i)}_t |s^{(i)}_{t-1}) \right],
\end{equation*}
where $c$ is a baseline for sample $i$ that is a function of the $i$th trajectory's state up to time $t-1$ as well as the returns produced by all other trajectories. The goal is to pick a $c$ that is a good estimate of the return, and a straightforward choice of $c$ is the average return from the other samples
\begin{equation*}
    c = \frac{1}{k-1} \sum_{j \neq i}R^{(j)}_{t}.
\end{equation*}
This ignores the fact that $s^{(i)}_{t} \neq s^{(j)}_{t}$, which can make this standard baseline unusable. For example, in our setting different trajectories may have emitted different numbers of tokens on a given timestep, resulting in substantial differences in return between trajectories that do not indicate the relative merit of those trajectories. Ideally, we would average over multiple trajectories starting from $s^{(i)}_{t}$, but this is computationally expensive. In \cite{variationalimco} the authors propose the following baseline which adds a residual term to address this. Let $r_t = \log p(y_{t}| s_{t}, b_{t})$ be the instantaneous reward at timestep $t$, then the baseline at timestep $t$ can be written 
\begin{equation}
c = \frac{1}{k-1} \sum_{j \neq i} R^{(j)}_{t} + \frac{1}{k-1} \sum_{j \neq i} \sum_{t' < t} r^{(j)}_{t'} - r^{(i)}_{t'}.
\label{eq:vimco_c}
\end{equation}
This baseline results in a learning signal that is the same across all timesteps, potentially increasing variance as all decisions in a trajectory are rewarded or punished together. We will call this the \emph{leave-one-out} (LOO) baseline because the baseline for a given sample is constructed using an average of the return of the other $k-1$ samples. Note that VIMCO optimizes the multisample variational lower bound in equation (\ref{eq:ms_vi_lowerbound}) with the leave-one-out baseline, and NVIL optimizes the single sample variational lower bound in equation (\ref{eq:ss_vi_lowerbound}) with a baseline that can be learned or computed from averages \cite{neuralvil}.

As the return strongly depends on the number of emitted tokens at time $t$, we can instead average the return of the other samples from when they have emitted the same number of tokens as sample $i$. In particular, let $e^{(j)}_t = \min_{t'} O^{(j)}(t') \geq O^{(i)}(t)$ be the first timestep when sample $j$ has emitted the same number of tokens as sample $i$ at timestep $t$, then
\begin{equation}
c = \frac{1}{k-1} \sum_{j \neq i} \sum_{t' > e_{t-1}^{(j)}}^T r^{(j)}_{t'}.
\label{eq:pastfuture_c}
\end{equation} 
We call this new baseline the \emph{temporal leave-one-out} baseline because it takes into account the temporal reward structure of our setting. This baseline can be combined with the parametric baseline, and is applicable to both variational inference and REINFORCE objectives in single- and multi-sample settings. We explore the performance of these baselines empirically in the experiments section.

\section{Related Work}
In this section we first highlight the relationship between our model and other models for attention.  Tang et. al. \cite{tang2014learning} proposed visual attention within the context of generative models, while Mnih et. al. \cite{mnih2014recurrent} proposed using recurrent models of visual attention for discriminative tasks. Subsequently, visual attention was used in an image captioning model \cite{xu2015show}. These forms of attention use discrete variables for attention location. Recently, `soft-attention' models were proposed for neural machine translation and speech recognition \cite{bahdanau,chorowski}. Unlike the earlier mentioned, hard-attention models, these models pay attention to the entire input and compute features by blending spatial features with an attention vector that is normalized over the entire input. Our paper is most similar to the hard attention models in that features at discrete locations are used to compute predictions. However it is different from the above models in the training method: While the hard attention models use REINFORCE for training, we follow variational techniques. We are also different from the above models in the specific application -- attention in our models is over temporal locations only, rather than visual and temporal locations. As a result, we additionally propose the temporal leave-one-out baseline.

Because the attention model we use is hard-attention, the model we use has parallels to prior work on online sequence-to-sequence models \cite{neural_transducer,yuping}. The neural transducer model \cite{neural_transducer} can use either hard attention, or a combination of hard attention with local soft attention.  However it explicitly splits the input sequences into chunks, and it is trained with an approximate maximum likelihood procedure that is similar to a policy search. The model of Luo et. al. \cite{yuping} is most similar to our model. Both models use the same architecture; however, while they use REINFORCE for training, we explore VIMCO for training the attention model. We also propose the novel temporal LOO baseline. A similar model with REINFORCE has also been used for training an online translation model \cite{learning-to-translate} and for training Neural Turing Machines \cite{rl-ntm}. Our work would be equally valid for these domains, which we leave for future work.

There has also been work using reweighted wake sleep to train sequential models. In \cite{ba2014multiple}, Ba et. al. optimize a variational lower bound with the prior instead of using a variational posterior. In this work, we refer to this as REINFORCE to distinguish it from variational inference with an inference network. In \cite{ba2015learning} the authors revisit this topic, using reweighted wake sleep to train similar models. Their algorithm makes use of an inference network but does not optimize a variational lower bound. Instead they optimize separate objectives for the model and the inference network that produce a biased estimate of the gradient of the log marginal likelihood.

\begin{figure*}[t]
\vspace{-9mm}
\begin{center}
\begin{minipage}{.32\linewidth}
    \includegraphics[width=1\linewidth]{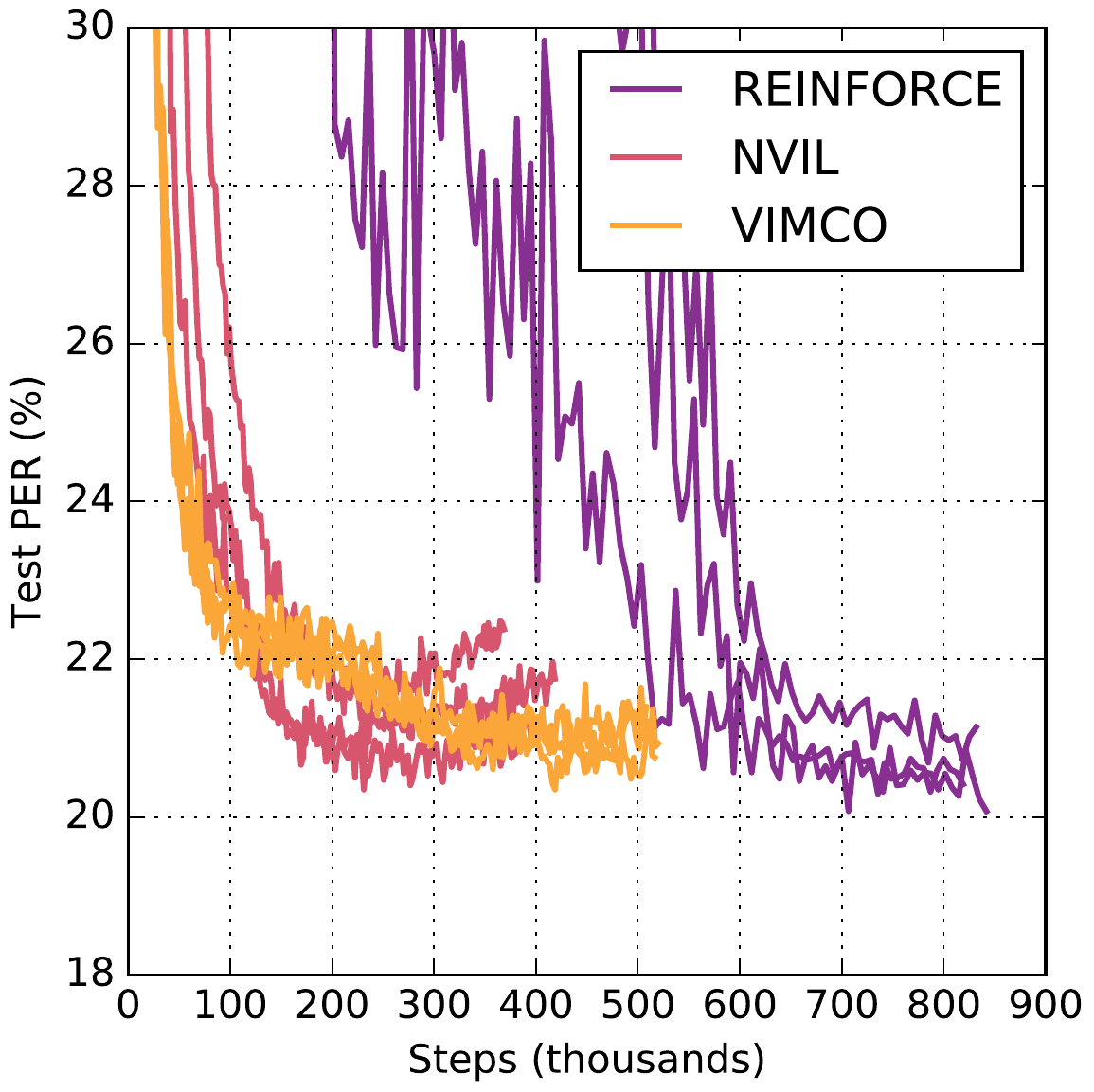}
\end{minipage}
\begin{minipage}{.3225\linewidth}
    \includegraphics[width=1\linewidth]{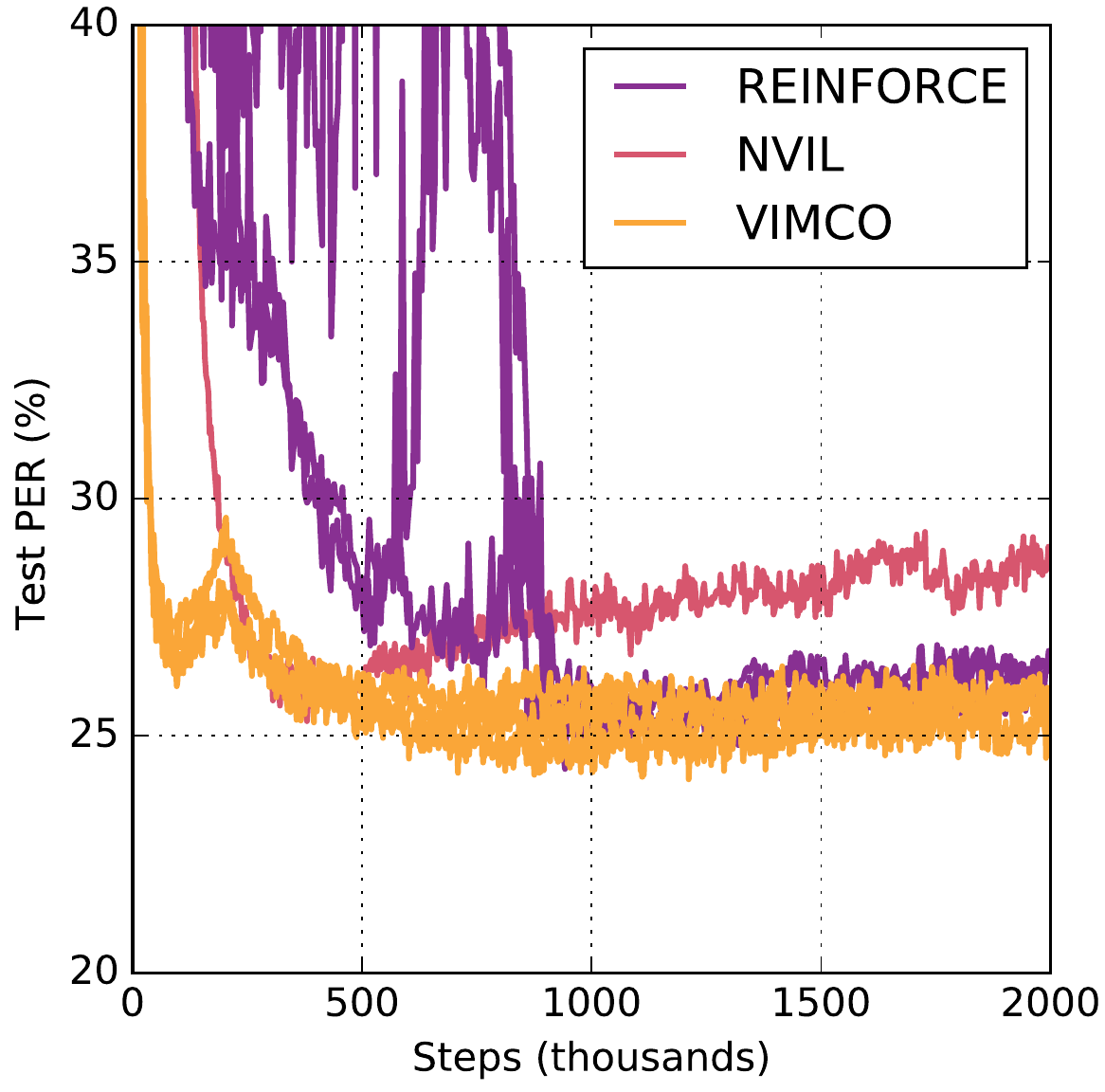}
\end{minipage}
\begin{minipage}{0.32\linewidth}
    \vspace{-4mm}
    \includegraphics[width=1\linewidth]{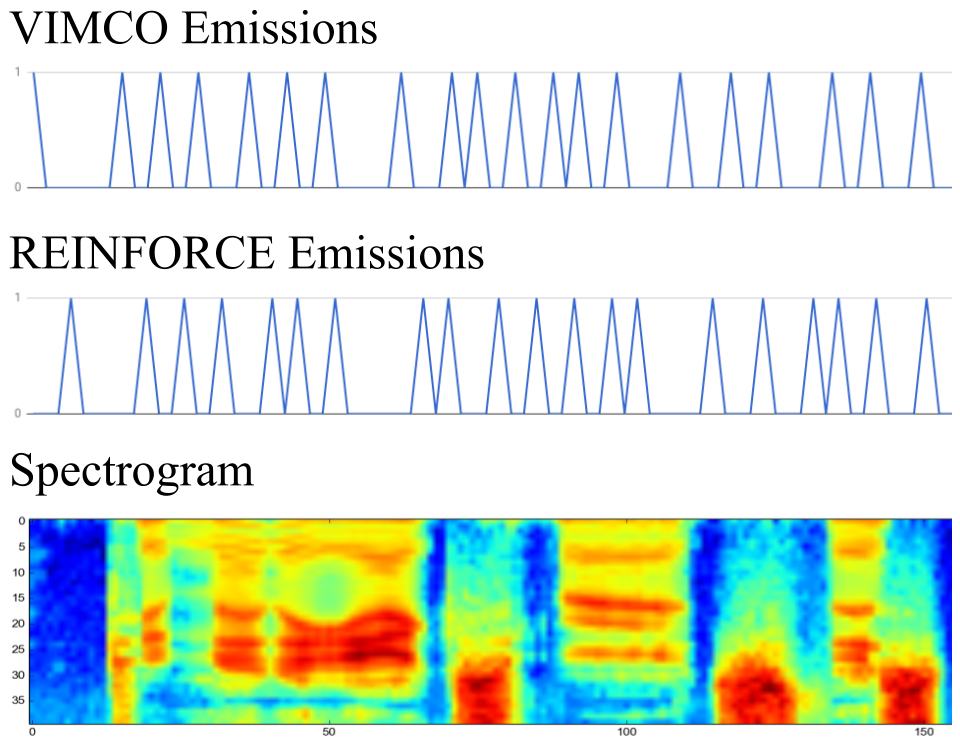}
\end{minipage}
\vspace{-3mm}
\caption{Test set phoneme error rate (PER) curves for models trained with REINFORCE, NVIL, and VIMCO on the TIMIT dataset (left), the Multi-TIMIT $10\%$ mixing proportion dataset (middle), and sample emission decisions for different methods on a TIMIT utterance (right). We evaluated three independent trials for each method. VIMCO converged more quickly than REINFORCE on both datasets. Furthermore, the performance gap between REINFORCE and VIMCO increases with Multi-TIMIT. We hypothesize that because Multi-TIMIT is a more challenging task, having a strong approximation to the posterior lets the model draw attention to the correct positions. NVIL performed well on TIMIT, but struggled with the more challenging Multi-TIMIT (note that only a single trial performs reasonably).}
\label{training}
\end{center}
\vspace{-10mm}
\end{figure*} 

\section{Experiments}
For our experiments we used the standard TIMIT phoneme recognition task. The TIMIT dataset has 3696 training utterances, 400 validation utterances, and 182 test utterances. The audio waveforms were processed into frames of log mel filterbank spectrograms every 25ms with a stride of 10ms. Each frame had 40 mel frequency channels and one energy channel; deltas and accelerations of the features were append to each frame. As a result each frame was a 123 dimensional input. The targets for each utterance were the sequence of phonemes. We used the 61 phoneme labels provided with TIMIT for training and decoding. To compute the phone error rate (PER) we collapsed the 61 phonemes to 39 as is standard on this task \cite{lee1989speaker}.

To model $p$ we used a 2-layer LSTM with 256 units in each layer. For the variational posterior $q$ we first processed the inputs $x_{1:T}$ with a 4-layer bidirectional LSTM and then fed the final layer's hidden state $h'$ into a 2-layer unidirectional LSTM along with the current target $y_t$ and the previous emission decision $b_{t-1}$. Each layer had 256 units. Note that in this case the approximate posterior does not have access to $y_{t+1:T}$ at timestep $t$ --- in practice we found giving $q$ access to $y$ far in the future did not improve performance. 

We regularized the models with variational noise \cite{gravesvariational} and performed a grid search over the values $\{0.075, 0.1, 0.15 \}$ for the standard deviation of the noise. We also used L2 regularization and grid searched over the values $\{1\times10^{-5}, 1 \times 10^{-4}, 1 \times 10^{-3} \}$ for the weight of the regularization.

\begin{table}[t]
  \caption{PER results on TIMIT test set for various models. This shows that REINFORCE performs comparably to the variational inference methods and that our novel baselines improve training for REINFORCE. It also shows that our baselines improve performance over \cite{yuping} which uses the same model with parametric baselines. Each number is the average of three runs. Our methods are above the horizontal line, while methods from the literature are listed below it. 
 }
  \vspace{0.2cm}
  \label{tab:timit}
  \centering
  \begin{tabular}{lccc}
    \hline
      Method & PER   \\
    \hline
    REINFORCE with leave-one-out (LOO) baseline & 20.5 \\
    NVIL with LOO baseline & 21.1 \\
    VIMCO with LOO baseline & \textbf{20.0} \\
    REINFORCE with temporal LOO baseline & \textbf{20.0} \\
    NVIL with temporal LOO baseline & 21.4 \\
    VIMCO with temporal LOO baseline & \textbf{20.0} \\
    \hline
    Online Alignment RNN (stacked LSTM) \cite{yuping} & 21.5 \\
    Neural Transducer with unsupervised alignments \cite{neural_transducer} & 20.8 \\
    Online Alignment RNN (grid LSTM) \cite{yuping} & 20.5 \\ 
    Monotonic Alignment Decoder \cite{raffel2017online}  & 20.4 \\
    Neural Transducer with supervised alignments \cite{neural_transducer} & 19.8 \\
    Connectionist Temporal Classification \cite{graves2006connectionist} & \textbf{19.6} \\
    \hline
  \end{tabular}
  \vskip -0.2in
\end{table}

\begin{table}[t]
  \caption{PER results on Multi-TIMIT for various algorithms. It can be seen that for this task VIMCO outperforms REINFORCE, and both VIMCO and REINFORCE outperforms RNN trained with Connectionist Temporal Classification significantly. The benefit of VIMCO increases as the second speaker's volume increases.}
  \label{tab:multitimit}
  \vspace{0.22cm}
  \centering
  \begin{tabular}{lccc}
  \hline
  \multirow{2}{*}{Method} & 
  \multicolumn{3}{c}{Mixing Proportion} \\
  & 0.50 & 0.25 & 0.1 \\
  \hline
      Connectionist Temporal Classification & 43.8 & 33.3 & 27.3 \\
      RNN Transducer & 48.9 & 32.2 & 25.7 \\
      REINFORCE with LOO baseline   & 42.9   & 32.5 & 25.9 \\
      NVIL with LOO baseline        & 70.1   & 71.8 & 55.2 \\
      VIMCO with LOO baseline       & \textbf{41.7}   & \textbf{30.7} & 25.4 \\
      REINFORCE with temporal LOO baseline   & 43.5 & 31.6 & 25.6 \\
      NVIL with temporal LOO baseline & 74.3 & 71.9 & 54.9 \\
      VIMCO with temporal LOO baseline & \textbf{41.7} & 30.75 & \textbf{25.2} \\
    \hline
  \end{tabular}
\end{table}

\subsection{Multi-TIMIT}
We generated a multi-speaker dataset by mixing male and female voices from TIMIT.  Each utterance in the original TIMIT dataset was paired with an utterance from the opposite gender. The waveform of both utterances was first scaled to lie within the same range, and then the scale of the second utterance was reduced to a smaller volume before mixing the two utterances.  We used three different scales for the second utterance: 50\%, 25\%, and 10\%.  The new raw utterances were processed in the same manner as the original TIMIT utterances, resulting in a 123 dimensional input per frame. The transcript of the speaker 1 was used as the ground truth transcript for this new utterance. Multi-TIMIT has the same number of train, dev, and test utterances as the original TIMIT, as well as the same target phonemes.

We trained models with the same configuration described above on the $3$ different mixing scales, and also trained 2-layer unidirectional LSTM models with Connectionist Temporal Classification for comparison. The results are shown in Table~\ref{tab:multitimit}.

\section{Results}
Figure~\ref{training} shows a plot of the training curves for the different methods of training and the different datasets. The variational methods (VIMCO and NVIL) require many fewer training steps compared to REINFORCE on both datasets. All methods used the same batch size and number of samples, so training steps are comparable. NVIL performs  well enough on a simple task like TIMIT, but struggles with Multi-TIMIT. It can be seen that the gap between REINFORCE and VIMCO increases on Multi-TIMIT (also see table~\ref{tab:multitimit}).

The right panel of Figure \ref{training} shows that REINFORCE attempts to wait to emit outputs until more information has come in, compared to VIMCO. This is presumably because it requires more information during learning. VIMCO, on the other hand, leverages the variational posterior which can access future $y$ and find the optimal place to emit.

In our experiments the difference between the performance of VIMCO and REINFORCE was larger for the more complicated task of Multi-TIMIT than for the simpler task of TIMIT. This can be explained by considering the samples that the models learn from. In the simpler problem of single speaker TIMIT, Monte-Carlo samples generated by REINFORCE have very high likelihood under $p(b|x)$ -- there are only a small number of samples that explain the entire probability mass, and these are sampled easily by a left to right ancestral pass (in time) of the model. These are very similar to the samples generated by the approximate posterior from VIMCO. As a result both methods perform approximately the same. In the case of Multi-TIMIT, however, in the ancestral pass the probabilities for individual emissions are much lower. Thus the likelihood is less 'peaked', and a large diversity of samples is chosen, leading to higher variance and poor learning. VIMCO, on the other hand does not face this problem because it samples from the approximate posterior, which is close to the true posterior and so very peaked around the `correct' samples of experience.

\section{Conclusion}
In this paper we have showed how we can adapt VIMCO to perform hard attention for the case of temporal problems and introduce a new variance-reducing baseline. Our method outperforms other methods of training online sequence to sequence models, and the improvements are greater for more difficult problems such as noisy mixed speech. In the future we will apply these  techniques to other challenging domains, such as visual attention.

\pagebreak

\bibliographystyle{IEEEbib}
\bibliography{paper}

\end{document}